\documentclass[letterpaper, 10 pt, conference]{ieeeconf}  

\IEEEoverridecommandlockouts                              

\usepackage[english]{babel}
\usepackage{amsfonts}
\usepackage[letterpaper,top=2cm,bottom=2cm,left=3cm,right=3cm,marginparwidth=1.75cm]{geometry}
\usepackage{amsmath}
\usepackage{graphicx}
\usepackage[customcolors]{hf-tikz}
\definecolor{perfblue}{RGB}{64, 114, 175}
\usepackage{cite} 

\usepackage{hyperref}
\hypersetup{
    colorlinks=true,
    citecolor=perfblue,
    linkcolor=perfblue,
    urlcolor=perfblue
}
\let\labelindent\relax
\usepackage{enumitem}
\usepackage{caption}
\usepackage{subcaption}
\usepackage[utf8]{inputenc}
\usepackage[linesnumbered,ruled,vlined]{algorithm2e}
\usepackage{amssymb}
\usepackage{setspace}
\usepackage{caption}
 \usepackage{algorithmic} 
 
\usepackage{xspace}
 \newcommand{\critq}[0]{\texttt{CritiQ}\xspace}
 \newcommand{\retry}[0]{\texttt{ReTRy}\xspace}
\usepackage{transparent}
\newcommand{\algcommentlight}[1]{\textcolor{perfblue}{\transparent{0.8}\small{\texttt{\textbf{//\hspace{2pt}#1}}}}}

\title{\LARGE \bf
Distilling Realizable Students from Unrealizable Teachers
}

\author{Yujin Kim$^{*, 1}$, Nathaniel Chin$^{*, 1}$, Arnav Vasudev$^{1}$, Sanjiban Choudhury$^{1}$
\thanks{$^{*}$ Equal contribution.}%
\thanks{$^{1}$Authors are with the Department of Computer Science, Cornell University, 14850, New York, United States. Emails: {yk826, nlc62, av432, sc2582}@cornell.edu}%
}

\begin{document}

\maketitle
\thispagestyle{empty}
\pagestyle{empty}

\begin{abstract}
We study policy distillation under privileged information, where a student policy with only partial observations must learn from a teacher with full-state access. A key challenge is \emph{information asymmetry}: the student cannot directly access the teacher’s state space, leading to distributional shifts and policy degradation.
Existing approaches either modify the teacher to produce realizable but sub-optimal demonstrations or rely on the student to explore missing information independently, both of which are inefficient. Our key insight is that the student should strategically interact with the teacher \textemdash querying only when necessary and resetting from recovery states \textemdash to stay on a recoverable path within its own observation space.
We introduce two methods: (i) an imitation learning approach that adaptively determines \emph{when} the student should query the teacher for corrections, and (ii) a reinforcement learning approach that selects \emph{where} to initialize training for efficient exploration.
We validate our methods in both simulated and real-world robotic tasks, demonstrating significant improvements over standard teacher-student baselines in training efficiency and final performance. The project website is available \href{https://portal-cornell.github.io/CritiQ_ReTRy/}{here}.
\end{abstract}

\section{Introduction}

Robots operating in the real world must learn to act effectively despite partial observations and limited ability to explore. Unlike in simulation, where policies have access to privileged state information, real-world policies must make decisions based on incomplete inputs~\cite{torne2024reconciling, hu2024privileged, hafner2023mastering}. A promising approach to this challenge is \emph{teacher-student policy distillation}, where a teacher policy—trained with full-state access—guides a student policy that operates under realistic sensing constraints. This method has been successfully applied in quadrupedal locomotion~\cite{miki2022learning, lee2020learning}, robotic manipulation~\cite{ankile2024imitation}, and visual navigation~\cite{uppal2024spin}.

A fundamental challenge in this setup is \emph{information asymmetry}: the teacher’s guidance is often unrealizable in the student’s observation space~\cite{warrington2021robust, messikommer2024student, walsman2022impossibly, weihs2021bridging}. Consider a robot searching for an item in a cluttered room. A teacher with privileged information knows the exact location of the item and follows a direct path to the object, while the student, with only partial visibility, cannot trace such a path. This leads to \emph{state aliasing}~\cite{bellemare2017distributional} where multiple teacher states collapse into the same student observation, resulting in conflicting teacher actions for the same input. Since the student cannot resolve these ambiguities, blindly imitating the teacher’s actions may lead to a trajectory that diverges from the intended solution, causing distributional shift and degraded performance.

Existing approaches manage information asymmetry by either adapting the teacher or exploring with the student. One line of work modifies the teacher’s policy to generate realizable but sub-optimal demonstrations~\cite{warrington2021robust, messikommer2024student}. Another line combines imitation learning with reinforcement learning (RL), allowing the student to explore and infer missing information~\cite{walsman2022impossibly, weihs2021bridging}. However, these methods suffer from fundamental trade-offs: constraining the teacher limits performance while blending IL and RL can lead to conflicting signals that destabilize training. 

Our key insight is that instead of blindly imitating the teacher, the student should \emph{strategically interact} with the teacher to remain on a recoverable trajectory. By querying the teacher only at critical moments and resetting to teacher recovery states, the student can optimize its policy to stay on a recoverable path within its own observation space.

We introduce two novel methods based on this insight.
(1) \emph{Critical State Query} (\critq): An imitation learning method that queries the teacher only in states where the student risks entering an unrecoverable trajectory.
(2) \emph{Resetting to Teacher Recovery} (\retry): A reinforcement learning method that resets the student to teacher recovery states to enable sample-efficient learning. 

Our contributions are:
\begin{enumerate}[nosep, leftmargin=0.2in]
\item We formalize teacher-student distillation under information mismatch and analyze the limitations of fundamental algorithms such as DAgger~\cite{ross2011reduction} in these settings.
\item We propose two novel IL and RL algorithms, \critq and \retry, that achieve improved performance bounds under state aliasing.
\item We empirically validate our approach in both simulated and real-world robotics environments, demonstrating significant improvements over standard teacher-student baselines.
\end{enumerate}

\begin{figure*}[hbt]
    \centering
    \includegraphics[width=1\linewidth]{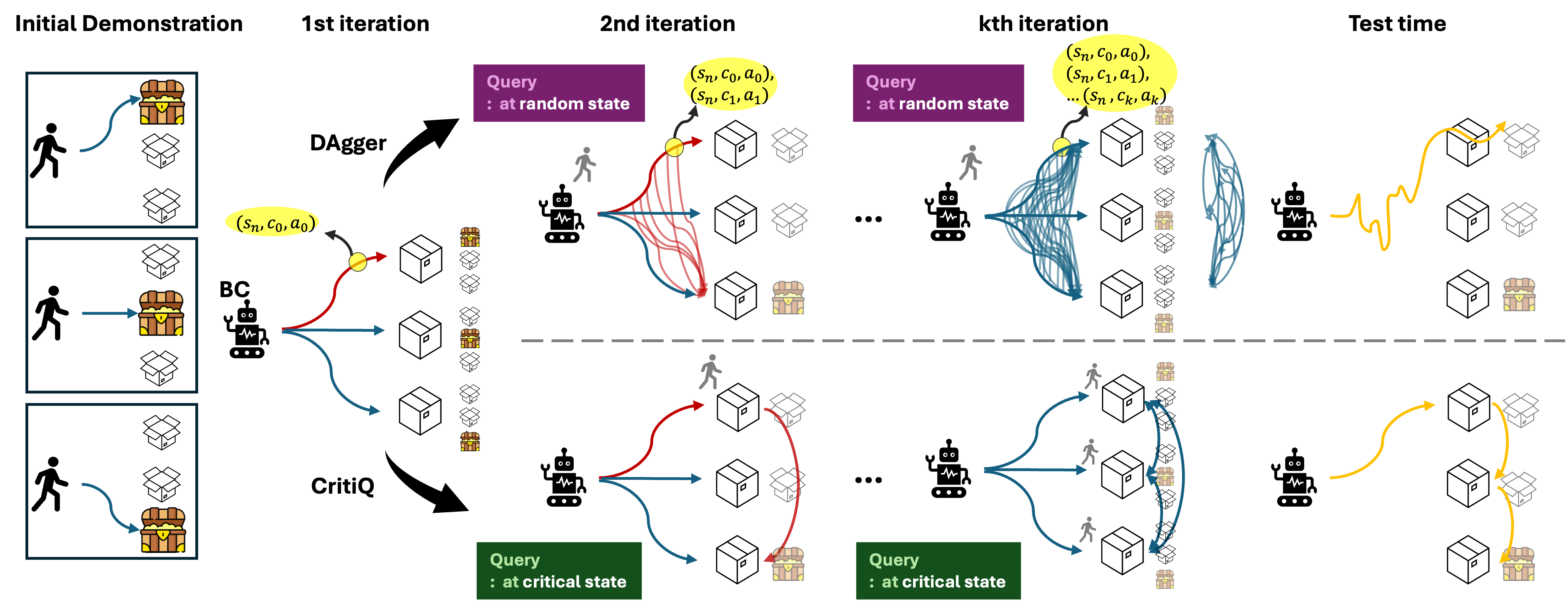}
    \caption{\textbf{DAgger vs \critq }.
    The goal is to find the box containing a coin, but the location is hidden from the student among three closed boxes. The student must search one-by-one. 
    In the first iteration, the student selects a trajectory from three demonstrations but cannot proceed further due to insufficient demontration. DAgger queries the expert at random states, leading to a dataset that becomes increasingly unrealizable, causing policy divergence. \critq, instead, queries only at critical states, ensuring necessary information is collected while maintaining realizability and improving policy stability.}
    \label{fig:CritiQ}
\end{figure*}

\section{Problem Formulation}

We formulate the problem as a Contextual Markov Decision Process (CMDP)~\cite{hallak2015contextual} defined by the tuple $(C, \mathcal{S}, \mathcal{A}, P, R, \gamma)$, where $\mathcal{S}$ is the state space, $\mathcal{A}$ is the action space, $P$ is the transition dynamics, $R$ is the reward function, $\gamma$ is the discount factor, and $C$ is the context, which affects both transitions and rewards. The context is privileged information available to the teacher but hidden from the student.

Training the student directly via reinforcement learning (RL) is intractable, as it requires solving a partially observable MDP (POMDP)~\cite{hallak2015contextual}. Instead, imitation learning (IL) algorithms such as DAgger~\cite{ross2011reduction} provide a promising alternative. Prior work~\cite{lee2020learning, ankile2024imitation, uppal2024spin} first trains a privileged teacher to solve the MDP, then interactively query this teacher to train a non-privileged student to imitate it. However, a fundamental challenge arises due to a mismatch between the teacher's and student's state representations.

Formally, we define the privileged teacher state as $\tilde{s} = (c, s)$, where the teacher has access to the full context $c$. The student state, in contrast, is given by a surjective mapping $f(\tilde{s}) = s$, meaning multiple teacher states $\tilde{s}$ may collapse to the same student state $s$. This state aliasing~\cite{bellemare2017distributional} prevents the student from distinguishing between different teacher-optimal actions. Even if the student perfectly predicts a teacher's action, it can be inconsistent across states leading to a highly suboptimal trajectory. We formalize this in Sec.~\ref{sec:critiq}, and show that under such aliasing, DAgger leads to poor performance. 

\section{Approach}
We propose a teacher-student distillation framework that learns under information mismatch. Our key insight is that instead of blindly imitating the teacher, the student should \emph{strategically interact} with the teacher to remain on a recoverable trajectory. We introduce two novel methods based on this insight.
(1) \emph{Critical State Query} (\critq): An imitation learning method that queries the teacher only in states where the student risks entering an unrecoverable trajectory. (Sec.~\ref{sec:critiq})
(2) \emph{Resetting to Teacher Recovery} (\retry): A reinforcement learning method that resets the student to teacher recovery states to enable sample-efficient learning. (Sec.~\ref{sec:retry})

\subsection{Critical State Query (\critq)}
\label{sec:critiq}

\subsubsection{Overview}
\label{sec:intuitive_critiq}

A key limitation of standard imitation learning methods like DAgger~\cite{ross2011reduction} is that they query the teacher for corrections at \emph{all states} visited by the student. While this improves error correction in standard MDPs, under state aliasing—where multiple teacher states collapse to the same student observation—this strategy worsens learning. Specifically, querying the teacher at every state generates conflicting labels for the same student observation, which destabilizes policy learning.

Consider the example in Fig.~\ref{fig:CritiQ}, where a student must find a box containing a gold coin. The teacher, with privileged access to the coin’s location, navigates directly to the correct box. However, the student, lacking this information, observes the teacher redirect the student to multiple different boxes for the same observation. If the student tries to follow such conflicting corrections, it learns a random dithering policy that fails to make systematic progress towards any one box. 

To mitigate this, we propose \critq  that queries the teacher only at critical states—states where the student is about to take an action with no recoverable supervision (e.g., choosing an incorrect box). This ensures that the student receives necessary correction data while avoiding excessive aliasing. By limiting queries to critical states, \critq enables the student to recover from errors without amplifying conflicting labels, leading to a structured search strategy rather than an unstable imitation of the teacher’s trajectory.

\subsubsection{Algorithm}

\begin{algorithm}[!t]
    \caption{\critq}
    \label{alg:critiq}
    \begin{algorithmic}[1]
        \STATE \textbf{Input:} Teacher policy $\pi^*$, discriminator threshold $\kappa$, iterations $N$
        \STATE \textbf{Output:} Trained student policy $\pi$
        \STATE Initialize $\pi_1 = \pi_{\rm BC}$, $G_{\phi}$, and $\mathcal{D}=\{\}$
        \FOR{$i = 1$ to $N$} 
            \STATE \algcommentlight{Roll out student policy t}
            \STATE Roll out $\pi_i$: $\tau_{\pi} = \{(s_t, a_t)\}_{t=0}^T$
            \FOR{each state $s_t$ in $\tau_{\pi}$}
                \STATE \algcommentlight{Check if state is critical using discriminator}
                \IF{$G_{\phi}(s_t) < \kappa$}
                    \STATE \algcommentlight{Query teacher}
                    \STATE Query teacher: $a^*_t = \pi^*(s_t)$
                    \STATE $\mathcal{D} \leftarrow \mathcal{D} \cup \{(s_t, a^*_t)\}$
                \ENDIF
            \ENDFOR
            \STATE \algcommentlight{Update student policy }
            \STATE Update $\pi_{i+1}$ to minimize $\mathcal{L}_S(\mathcal{D}, G_\phi)$
            \STATE \algcommentlight{Update discriminator }
            \STATE Update $G_{\phi}$ to minimize $\mathcal{L}_G(\pi_{i}, \pi^*)$
        \ENDFOR
        \RETURN Best policy $\pi \in \pi_{1:N+1}$ on validation
    \end{algorithmic}
\end{algorithm}

\critq selectively queries the teacher only at \emph{critical states}—states that the student visits that the expert has not visited. This prevents adding states that increasing state aliasing while ensuring the student receives enough recovery data to learn effectively.

To detect critical states efficiently, we train a discriminator $G_{\phi}$ to distinguish between teacher and student trajectories. The discriminator is trained using the loss 
\begin{equation}
\begin{aligned}
    \mathcal{L}_G (\pi, \pi^*)= &- \mathbb{E}_{\tau \sim \pi} [\log G_{\phi}(\tau)] \\
    &- \mathbb{E}_{\tau \sim \pi^*} [\log (1 - G_{\phi}(\tau))].
\end{aligned}
\end{equation}
where $\tau\sim\pi$, $\tau\sim\pi^*$ denote trajectories generated by the teacher and student, respectively. The student is simultaneously trained to minimize imitation loss on aggregated data $\ell(s, \mathcal{D})$ while fooling the discriminator $G_\phi$:
\begin{equation}
    \mathcal{L}_S(\mathcal{D}, G_\phi) = \ell(\mathcal{D}, \pi) - \lambda \mathbb{E}_{\tau \sim \pi}[G_{\phi}(\tau)].
\end{equation}
where $\lambda$ is a weighting factor balancing imitation and discriminator feedback.

Algorithm~\ref{alg:critiq} describes the iterative learning process for \critq. The student policy is initialized with behavior cloning. In each iteration, the student rolls out its policy to collect a trajectory of visited states. For each state encountered, a discriminator is used to determine whether the state is \textit{critical}—a state the expert has not previously visited. If the discriminator score falls below a threshold, the teacher is queried for the correct action, and the new state-action pair is added to the dataset.

After collecting critical corrections, the student policy is updated to minimize imitation loss while simultaneously training the discriminator to better differentiate between teacher and student trajectories. The process repeats for a fixed number of iterations, gradually refining both the student policy and the discriminator. At the end of training, the best-performing student policy is selected based on validation performance. This ensures that \critq effectively balances exploration and expert guidance, preventing excessive aliasing while still providing sufficient recovery data.
\subsubsection{Analysis}
In an MDP, the performance bound of DAgger~\cite{ross2011reduction} is:
\begin{equation}
J(\pi) \leq J(\pi^*) + u T \epsilon_N + O(1),
\end{equation}
where $\pi$ is the student policy, $\pi^*$ the teacher, $T$ is the episode length, $u$ is the advantage bound, and $\epsilon_N$ is the best achievable imitation loss after $N$ iterations of data aggregation.

However, in the contextual MDP setting, where the student lacks privileged context $c$, the imitation loss decomposes into two terms:
\begin{equation}
\epsilon = \epsilon_{\rm model} + \delta.
\end{equation}
Here, $\epsilon_{\rm model}$ is the model capacity error from function approximation, and $\delta$ is the realizability error caused by state aliasing. Under DAgger, repeated querying introduces additional state-aliasing by aggregating conflicting expert actions at the same student state, causing $\delta$ to increase monotonically. Over multiple iterations, the dataset at a given student state $s_n$ grows as:
\begin{equation}
\mathcal{D}(s_n) = {(s_n, c_0, a_0), (s_n, c_1, a_1), \dots, (s_n, c_k, a_k)},
\end{equation}
where different contexts $c_0, c_1, ..., c_k$ map to different expert actions $a_0, a_1, ..., a_k$. Since the student cannot observe the context $c$, it faces inconsistent supervision at $s_n$, making it impossible to minimize imitation loss beyond a lower bound given by $\delta$. On the other hand, \critq limits teacher queries to states with no recovery data, ensuring that $\delta$ does not grow unnecessarily.

\begin{figure*}[!t]
    \centering
    \includegraphics[width=1\linewidth]{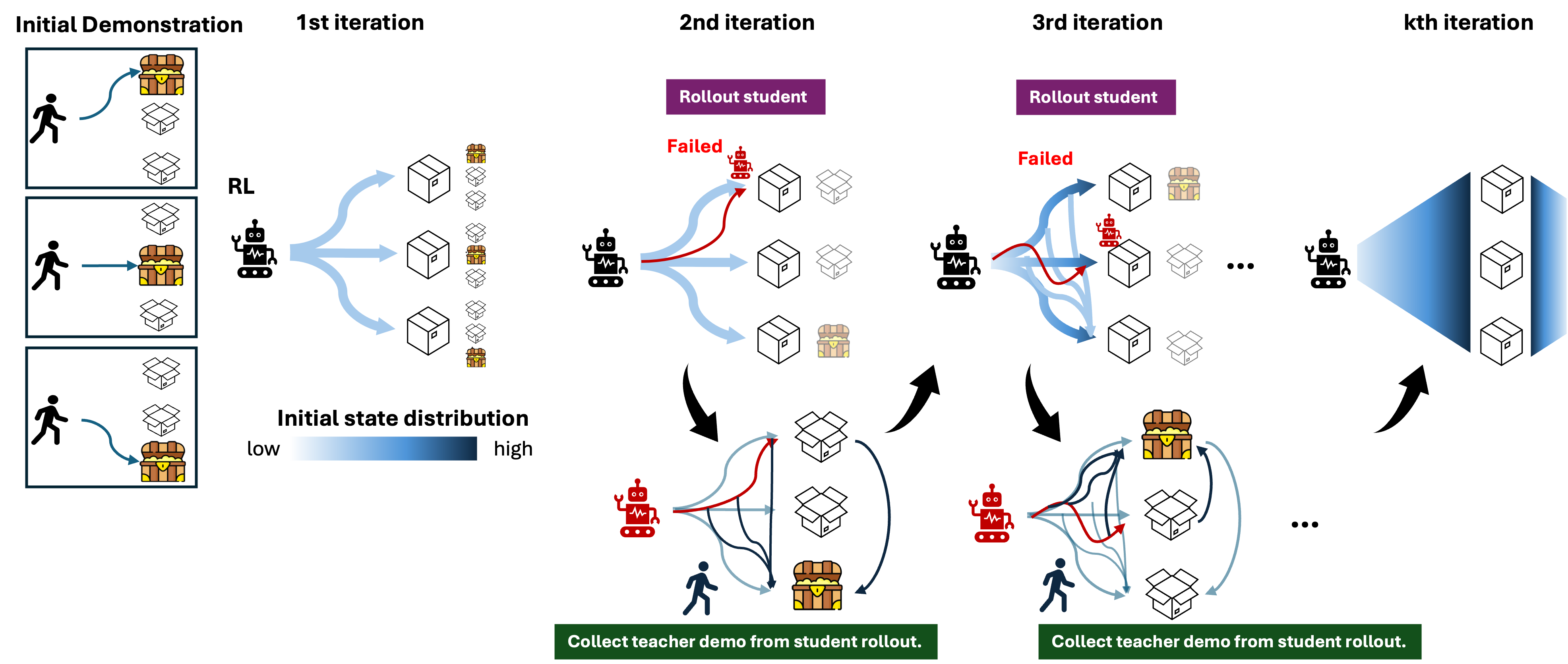}
    \caption{\textbf{\retry}.
    The bidirectional interaction between the teacher and student provides a feasible reset distribution, enabling the RL agent to acquire policies more effectively. As iterations progress, the RL student’s exploration space becomes denser, forming a structured search area for policy learning. The dark blue region represents high-probability reset states, while lighter blue areas indicate lower probability, and white regions have zero probability. }
    \label{fig:ResT}
\end{figure*}

Under DAgger, the number of expert demonstrations at state $s$ increases by batch size $B$ at every iteration $|\mathcal{D}_t(s)| = |\mathcal{D}_{t-1}(s)| + B$. Since each expert query introduces a new demonstration potentially corresponding to a different hidden context, conflicting labels accumulate. The realizability error grows as:
\begin{equation}
    \delta_t(s) \geq \delta_{t-1}(s) + \frac{1}{|\mathcal{D}_t(s)|} \sum_{i=1}^{B} \mathbb{I}[s \in \mathcal{S}_{\rm alias}].
\end{equation}
where $\mathcal{S}_{\rm alias}$ are the set of aliased states. Since this term is strictly positive in the presence of aliasing, the unrealizability error monotonically increases under DAgger.

In contrast, \critq limits teacher queries to critical states, where the student has no prior recovery data. Let $\mathcal{S}_{\text{critical}}$ be the set of critical states. The dataset expansion follows:
\begin{equation}
    |\mathcal{D}_t(s)| \approx |\mathcal{D}_{t-1}(s)|, \quad \forall s \notin \mathcal{S}_{\text{critical}}.
\end{equation}
This results in a much slower growth of realizability error $\delta_t$ compared to DAgger. 
See Appendix~\ref{app:critical_state} for full analysis.

\subsection{Resetting to Teacher Recovery (\retry)}
\label{sec:retry}
\label{sec:TCR}

\subsubsection{Overview}
RL is immune from the aliasing problems since the student disregards the teacher completely and figures out the optimal policy from trial and error. However, training RL agents under partial observability is challenging because the student must explore without access to privileged teacher information. A common strategy to accelerate RL is \emph{resets}~\cite{kakade2002approximately}, where the RL agent is initialized from states visited by a good exploration policy~\cite{florensa2017reverse, salimans2018learning}. This approach ensures that training starts from meaningful states where recovery is possible. However, teacher-based resets suffer from a key limitation: if the student encounters a state that was never visited by the teacher, it must rely on its own exploration, limiting the effectiveness of fixed resets.

\retry addresses this issue by iteratively refining the reset distribution. Instead of resetting only to teacher-visited states, \retry rolls out the teacher from states visited by the student, augmenting the set of reset states. This ensures that resets cover states that the best student policy is likely to visit. Unlike static teacher-based resets, \retry dynamically expands the set of reset states to improve exploration efficiency.

\subsubsection{Algorithm}

\begin{algorithm}[!t]
    \caption{\retry}
    \label{alg:teacher_correct_reset_rl}
    \begin{algorithmic}[1]
        \STATE \textbf{Input:} Teacher policy $\pi^*$,  iterations $N$
        \STATE \textbf{Output:} Trained student policy $\pi$
        \STATE Initialize $\pi$, $\mathcal{D}^{\pi^*}$, $\mathcal{D}^{\pi} = \{\}$
        \FOR{$i = 1$ to $N$} 
            \STATE \algcommentlight{Train student policy via RL with resetting to teacher states}
            \STATE Sample $s_0 \sim \mathcal{D}^{\pi^*}$
            \STATE Roll out $\pi_i$ from $s_0$: $\tau_{\pi} = \{(s_t, a_t)\}_{t=0}^T$
            \STATE Aggregate student states $\mathcal{D}^{\pi} \leftarrow \mathcal{D}^{\pi} \cup \tau^\pi$
            \STATE Update student $\pi$ via policy gradient: $\pi_{i+1} \leftarrow \pi_i + \alpha \nabla J(\pi)$
            \STATE \algcommentlight{Collect teacher rollouts from student-visited states}
            \STATE Sample $s_0 \sim \mathcal{D}^{\pi}$
            \STATE Roll out teacher $\pi^*$ from $s_0$: $\tau_{\pi^*} = \{(s_t, a_t)\}_{t=0}^T$
            \STATE $\mathcal{D}^{\pi^*} \leftarrow \mathcal{D}^{\pi^*} \cup \tau^{\pi^*}$
        \ENDFOR
        \RETURN Best policy $\pi \in \pi_{1:N+1}$ on validation
    \end{algorithmic}
\end{algorithm}

Algorithm~\ref{alg:teacher_correct_reset_rl} presents the \retry algorithm, which iteratively refines the student’s reset distribution to improve RL efficiency. The training process alternates between student rollouts and teacher rollouts, progressively expanding the set of reset states. At the beginning of each iteration, the student is reset to states previously visited by the teacher and collects experience by rolling out its policy. These rollouts are stored in the student-visited state dataset. The student’s policy is then updated using standard RL by applying policy gradient updates based on the collected data.

After the student has completed its rollouts, the teacher is rolled out from states visited by the student to generate new recovery demonstrations. The teacher’s rollouts expand the reset dataset to include additional recovery trajectories, ensuring that the student receives guidance even in states where the teacher was not originally present. These new demonstrations are added to the teacher-visited state dataset.

\subsubsection{Analysis}
The Performance Difference Lemma (PDL)~\cite{kakade2002approximately} states that the performance gap between an optimal realizable policy $\pi_{\rm opt}$ and a student policy $\pi$ is given by:
\begin{equation}
    J(\pi_{\rm opt}) - J(\pi) = \frac{1}{1 - \gamma} 
    \mathbb{E}_{\substack{s \sim d^{\pi_{\rm opt}}(s) \\ a \sim \pi(a|s)}} 
    \left[A^{\pi_{\rm opt}}(s, a)\right].
\end{equation}

where $d^{\pi_{\rm opt}}(s)$ represents the state visitation distribution of the optimal realizable policy, and $A^{\pi_{\rm opt}}(s, a)$ is the advantage function. Hence, efficient learning occurs when the reset distribution $d(s) \approx d^{\pi_{\rm opt}}(s)$. Since $d^{\pi_{\rm opt}}(s)$ is unknown, standard RL methods initialize the student from an arbitrary reset distribution $d(s)$, with the aim of bounding the density ratio $\left\| \frac{d(s)}{d^{\pi^*}(s)} \right\|_\infty \leq C$, which yields a performance bound of $J(\pi_{\rm opt}) - J(\pi) \leq C T \epsilon$.  A smaller $C$ leads to tighter performance bounds and faster convergence.

Resetting the student to teacher-visited states is a common strategy for improving sample efficiency~\cite{florensa2017reverse, salimans2018learning}. However, in CMDPs, the teacher’s reset distribution $d^{\pi^*}(s)$ is not realizable for the student due to state aliasing. If the density ratio $\left\| d^{\pi^*}(s) / d^{\pi_{\rm opt}}(s) \right\|_\infty$ is unbounded, direct resets to teacher states lead to slow convergence.

\retry mitigates this issue by iteratively refining the reset distribution. Initially, the student is reset only to teacher-visited states. As training progresses, the teacher is rolled out from student-visited states, gradually expanding the reset distribution. By progressively blending teacher and student rollouts, \retry ensures that the density ratio remains bounded. 

\section{Experiments}
\subsection{Setup}
We evaluate \critq and \retry on a set of manipulation and navigation tasks for a Hello Robot Stretch mobile manipulator. The simulated tasks are executed in Mujoco, while the real robot tasks are executed by training a teacher and student in sim and transferring to real. We evaluate on the following tasks:
\begin{figure}[t!]
    \centering
        \subfloat[\centering {Drawer sim}]{
        \includegraphics[width=.17\textwidth]{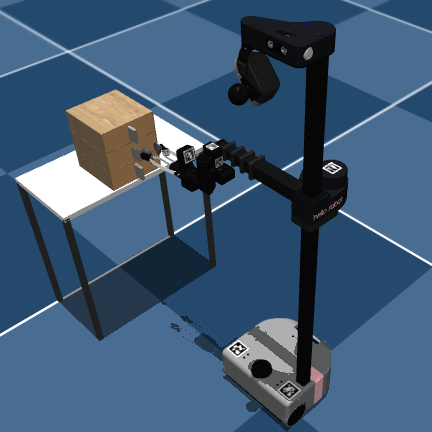}
        \label{fig:drawer}}
        \subfloat[\centering {Drawer real}]{
        \includegraphics[width=.17\textwidth, height=.17\textwidth]{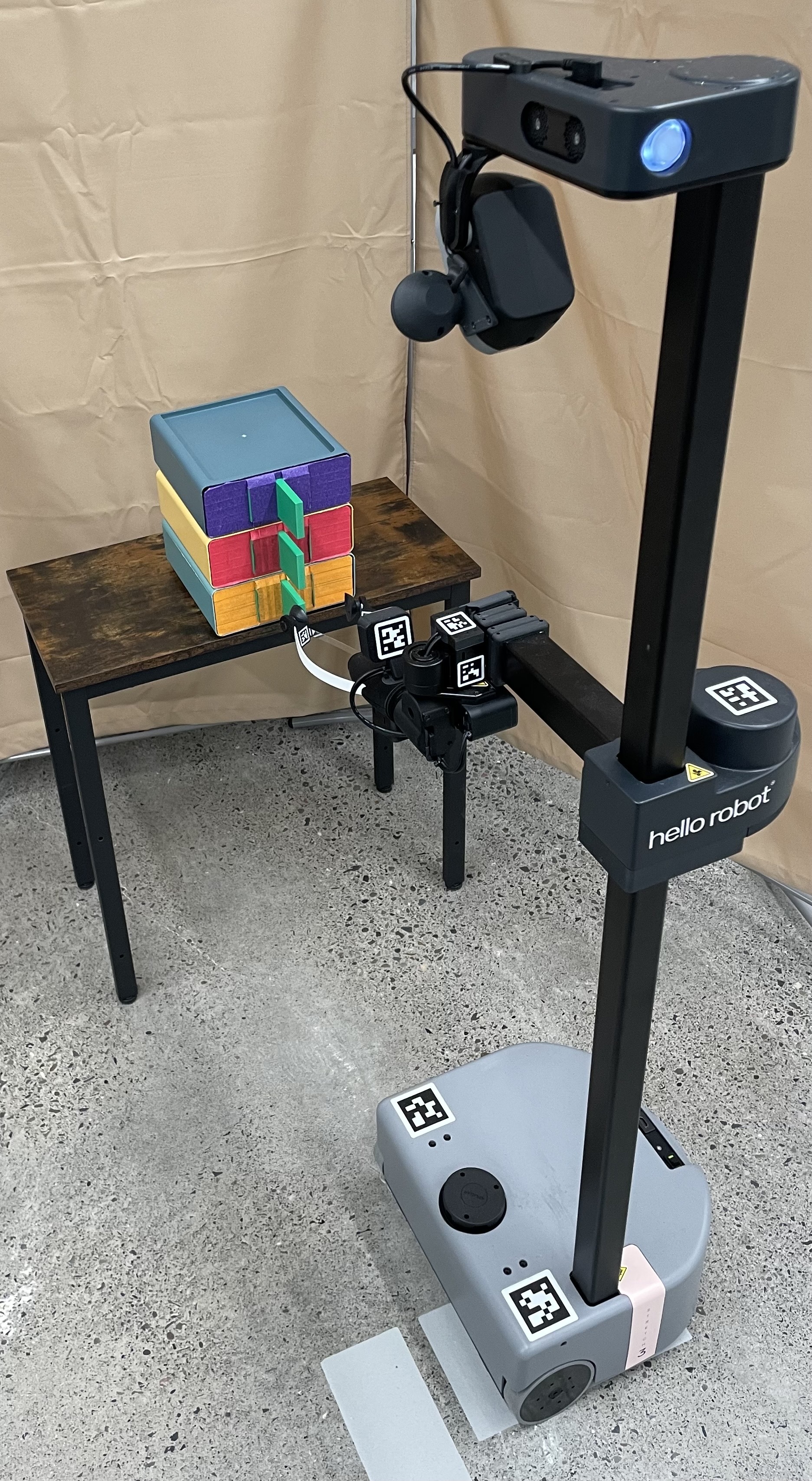}
        \label{fig:drawer_real}}
        \vfill
        \subfloat[\centering {Block push}]{
        \includegraphics[width=.17\textwidth, height=.17\textwidth]{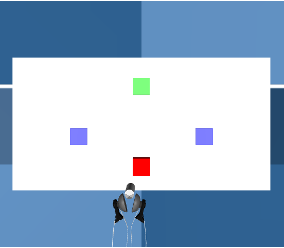}
        \label{fig:block}
        }
        \subfloat[\centering {Navigation}]{
        \includegraphics[width=.17\textwidth]{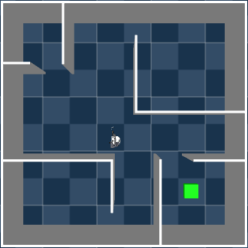}
        \label{fig:nav}
        }
    \caption{\label{fig:env}{\textbf{Simulation and real robot tasks}. In each of these tasks, the teacher has access to the goal, while the student needs to explore the different possible goals. }}
\end{figure}
\begin{enumerate}[nosep, leftmargin=0.2in]
    \item \textbf{Drawer Search (sim + real)}: Find the object that is in one of the three drawers. The teacher policy has access to which drawer contains the object, while the student only keeps track of which drawers have been opened throughout the episode. 

    \item \textbf{Push Block Search (sim)}: Push the block to the correct goal location out of 3 possible locations. The teacher policy has access to the correct goal location, while the student only keeps track of which goal locations the block has visited.  

    \item \textbf{Navigation Search (sim)}: Navigate to the room that contains an object out of 4 possible rooms. The teacher policy has access to which room the object is in, while the student policy only keeps track of which rooms the robot has visited. 
\end{enumerate}

We compare our method against imitation learning baselines (BC and DAgger) as well as the RL baseline, SAC. For BC, we collect a dataset of teacher trajectories to train a student policy. We then initialize DAgger training with the BC policy and aggregate teacher corrections on student rollouts to the dataset throughout the training process. We use the success rate as the evaluation metric. 

\subsection{Results and Discussions}
\begin{figure}[!t]
    \centering
    \includegraphics[width=\columnwidth]{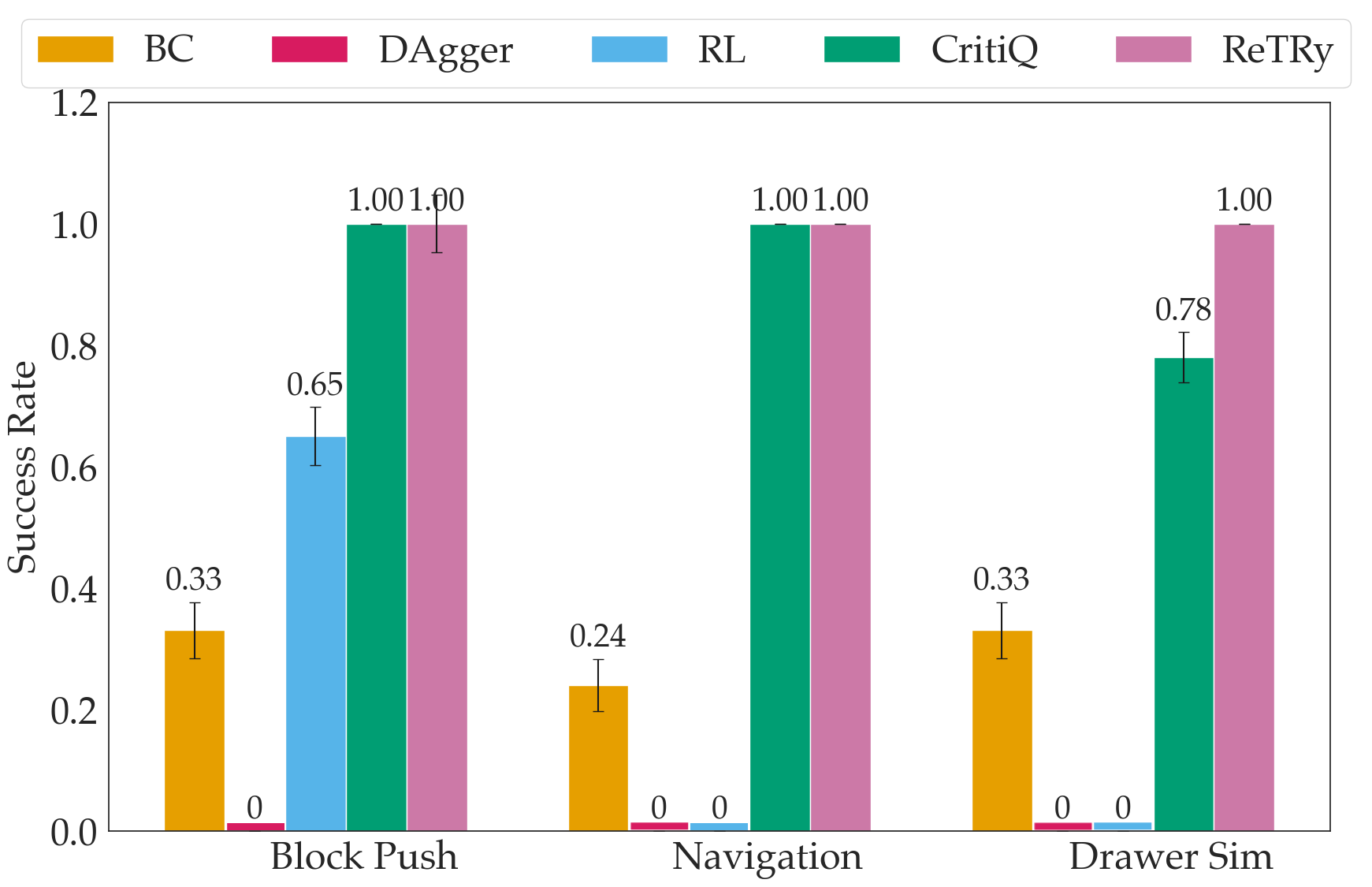}
    \caption{\textbf{Results on Simulation Tasks} We evaluate the success rate on 3 different robot tasks in simulation. We collected 100 episodes per method for all simulation tasks. Goals for each episode were sampled uniformly on a fixed seed.}
    \label{fig:overall_results}
\end{figure}

\begin{figure}[!t]
    \centering
    \includegraphics[width=\columnwidth]{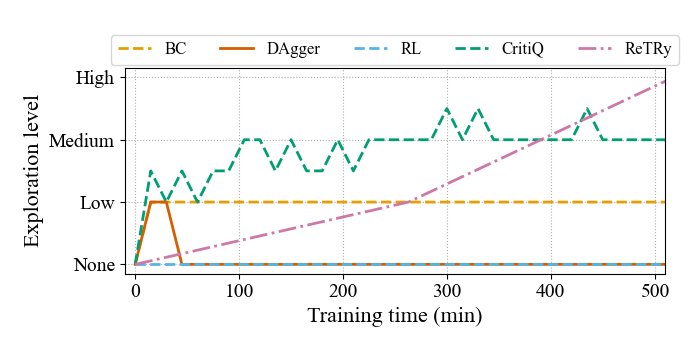}
    \caption{\textbf{Exploration level in the Drawer Search Task over Training Time.} If the student fails to explore any drawer, the exploration level is None. Once all three drawers are explored, the exploration level is considered High. The x-axis represents training time (minutes)}
    \label{fig:traintime}
\end{figure}
\textbf{Q1. Can student policies learn effectively from unrealizable teachers?}
Across all three simulated tasks (Fig. \ref{fig:overall_results}), \critq and \retry consistently outperform baselines. BC performs near random, selecting one teacher trajectory without recovery. DAgger fails entirely due to accumulating conflicting labels from the aliased teacher states. RL achieved a 0\% success rate on all tasks except the block-pushing task, where it reached 65\%, which has a relatively smaller exploration space.
In contrast, \retry achieves near-perfect success through the teacher state reset strategy, while \critq succeeds in most cases by querying selectively—except in the more challenging drawer task, where its success rate drops due to its reliance on behavior cloning.

\textbf{Q2. Do \critq and \retry enable effective exploration?} 
We introduce the exploration level metric to quantify how thoroughly agents explore possible goal options. As shown in Fig. \ref{fig:traintime}, RL struggles to explore and ultimately converges to a no-exploration level, as the sparse reward function does not incentivize exploratory behavior. In contrast, both \critq and \retry improve exploration over the course of training. \retry eventually achieves full exploration but requires longer training time, while \critq improves more rapidly in the early stages but saturates at a mid-level of exploration.\\
\begin{figure}[!t]
    \centering
    \includegraphics[width=\columnwidth]{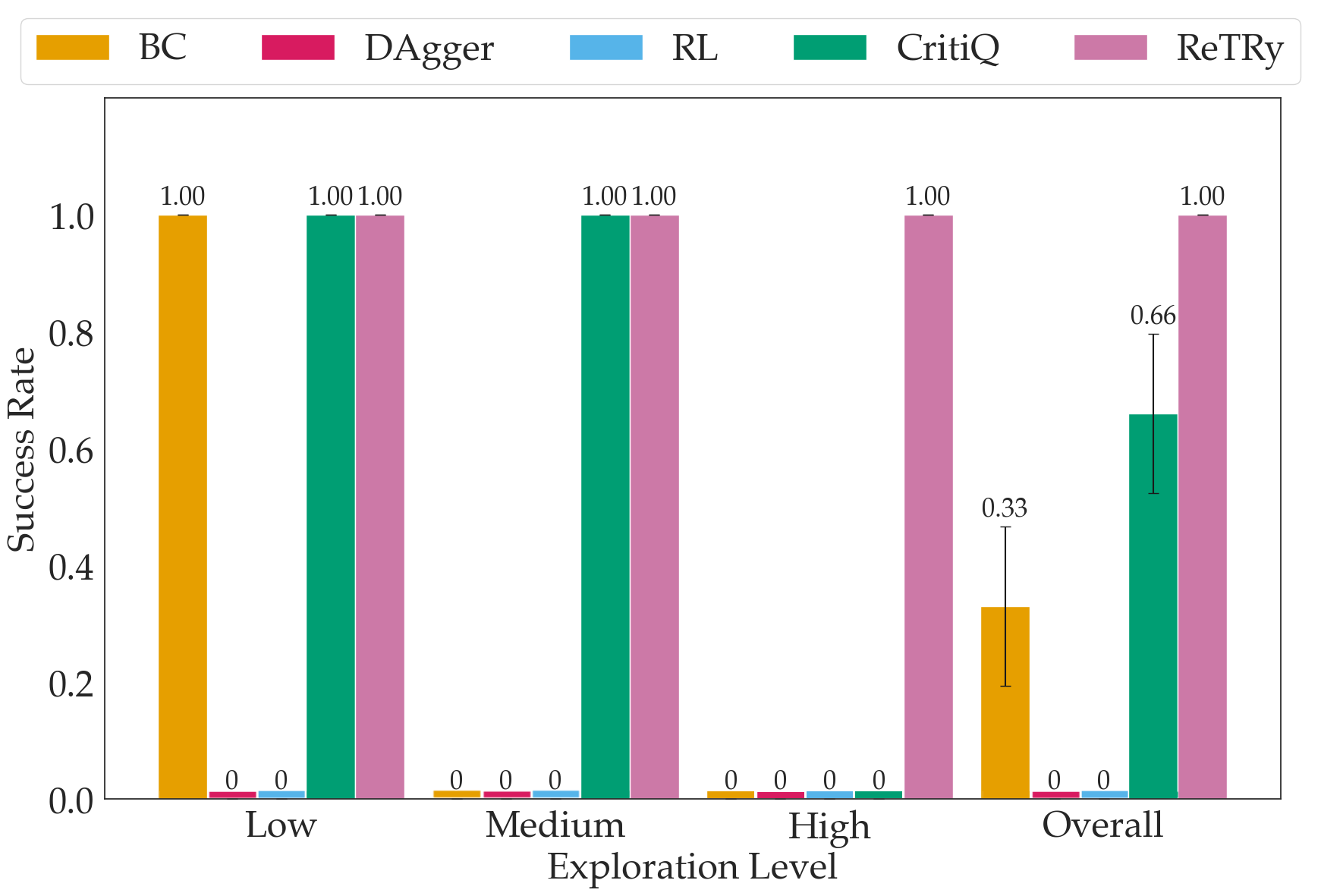}
    \caption{\textbf{Results on Drawer Task with Real Robot} We evaluate the success rate for each method on the drawer task with a real robot. We categorize exploration in hindsight as low, medium, and hard to indicate that the robot was able to explore one, two, or three drawers during the episode when necessary to do so. We collected 12 episodes per method and also recorded the overall success.}
    \label{fig:real_robot}
\end{figure}
\textbf{Q3. Do these policies transfer to real-world settings?}
In the real-world drawer search task (Fig. 6), \retry maintains a 100\% success rate, consistent with its simulation performance. In contrast, \critq sees a performance drop from 78\% in simulation to 66\% in reality, likely due to a distribution shift from sim to real caused by subtle differences in drawer dynamics. This shift leads to error accumulation from cloned policies and limits \critq’s ability to make effective recovery actions, particularly after opening two drawers. Nonetheless, it still outperforms baseline methods, which fail to generalize to high-exploration regimes.

\textbf{Q4. Why does \retry outperform \critq?}
\retry’s robustness stems from its reinforcement learning foundation and its adaptive reset strategy. It avoids compounding errors and learns from sparse rewards and structured exploration. \critq, in contrast, is susceptible to two issues: (1) early-stage discriminator errors can misclassify critical states, introducing noise into the student’s training set, and (2) behavior cloning’s reliance on static feedback makes it fragile in long-horizon, high-variance tasks like drawer manipulation.

\textbf{Q5. When should we use \critq or \retry?}
Both methods address information asymmetry but suit different scenarios. \critq is effective when the teacher is near-optimal and training time is limited, especially in high-dimensional spaces where data efficiency matters. \retry is more robust in long-horizon tasks and tolerates suboptimal teachers, thanks to its RL foundation that enables better generalization and recovery. However, it requires longer training and can struggle in large exploration spaces, which limits scalability.\\
\section{Related Work}
Teacher-student policy distillation has been widely adopted across various domains, including autonomous driving \cite{chen2020learning}, robotics \cite{chen2023visual,torne2024reconciling, miki2022learning, uppal2024spin, qi2023hand, liu2024visual}, and large language models (LLMs) \cite{choudhury2024better}, to transfer knowledge from a more capable teacher model to a more efficient student model.

A major challenge in teacher-student policy distillation is overcoming realizability issues arising from asymmetric information. Many approaches model the student’s learning as a Partially Observable Markov Decision Process (POMDP) \cite{kaelbling1998planning}, assuming the student can infer missing information from history \cite{torne2024reconciling, miki2022learning, uppal2024spin, qi2023hand, liu2024visual}. However, privileged information is often irrecoverable, making direct imitation infeasible.

Contextual Markov Decision Processes (CMDPs) \cite{hallak2015contextual, swamy2022sequence} provide a more suitable framework by modeling hidden privileged information as an unobserved context variable. In this setting, standard imitation learning approaches such as BC or DAgger fail due to unrealizable teacher demonstrations \cite{warrington2021robust}.
Existing solutions fall into two categories: (1) modifying the teacher’s policy to match the student’s observability, or (2) combining imitation with exploration. The first approach regularizes demonstrations to fit what the student can observe—e.g., Messikommer et al. \cite{messikommer2024student} penalize actions hard to imitate, while Warrington et al. \cite{warrington2021robust} iteratively adapt the teacher policy. However, such modifications reduce optimality, as the teacher must deviate from its own policy.

Alternatively, keeping the teacher’s policy fixed while enabling the student to learn through both imitation and exploration offers greater flexibility. Weihs et al. \cite{weihs2021bridging} balance imitation loss and RL objectives via adaptive weighting, allowing the student to disregard suboptimal teacher actions. Walsman et al. \cite{walsman2022impossibly} train both a follower and an explorer policy, blending IL and RL for long-term success. However, conflicting imitation and reward signals can destabilize training.

Rather than modifying the teacher, structured student learning can improve stability. Under a hidden context, excessive expert corrections can lead to confusion, making intervention learning \cite{spencer2020learning} more effective than DAgger. Similarly, in RL, resetting the agent from demonstrated states improves exploration \cite{florensa2017reverse, salimans2018learning, tavakoli2018exploring, khandate2023sampling}, but prior methods assume no information gap between the teacher and student. In asymmetric settings, demonstrated resets may provide a useful initialization but require adaptation to account for information mismatch.

\section{Conclusion}
We introduced \critq and \retry, two novel teacher-student policy distillation methods designed to address information asymmetry in Contextual MDPs. \critq mitigates state aliasing by selectively querying the teacher at critical states, ensuring that the student receives useful corrections without exacerbating unrealizability errors. \retry improves sample efficiency in reinforcement learning by iteratively refining the reset distribution, enabling the student to explore recoverable states while benefiting from expert guidance.

Empirical results demonstrate that both methods significantly outperform standard policy distillation approaches across benchmark tasks. However, \critq's effectiveness depends on the accuracy of critical state detection, where an overly strong discriminator increases aliasing, and a weak discriminator leads to off-distribution queries. Additionally, \critq remains susceptible to compounding errors inherent in behavior cloning, explaining its performance gap relative to \retry.

Future work will explore principled methods for critical state detection, potentially reducing reliance on learned discriminators. Extending these techniques to broader imitation learning and reinforcement learning settings may further improve policy optimization under partial observability.





\bibliographystyle{IEEEtran}

\bibliography{reference}

\begin{thebibliography}{10}
\providecommand{\url}[1]{#1}
\csname url@samestyle\endcsname
\providecommand{\newblock}{\relax}
\providecommand{\bibinfo}[2]{#2}
\providecommand{\BIBentrySTDinterwordspacing}{\spaceskip=0pt\relax}
\providecommand{\BIBentryALTinterwordstretchfactor}{4}
\providecommand{\BIBentryALTinterwordspacing}{\spaceskip=\fontdimen2\font plus
\BIBentryALTinterwordstretchfactor\fontdimen3\font minus \fontdimen4\font\relax}
\providecommand{\BIBforeignlanguage}[2]{{%
\expandafter\ifx\csname l@#1\endcsname\relax
\typeout{** WARNING: IEEEtran.bst: No hyphenation pattern has been}%
\typeout{** loaded for the language `#1'. Using the pattern for}%
\typeout{** the default language instead.}%
\else
\language=\csname l@#1\endcsname
\fi
#2}}
\providecommand{\BIBdecl}{\relax}
\BIBdecl

\bibitem{torne2024reconciling}
M.~Torne, A.~Simeonov, Z.~Li, A.~Chan, T.~Chen, A.~Gupta, and P.~Agrawal, ``Reconciling reality through simulation: A real-to-sim-to-real approach for robust manipulation,'' \emph{arXiv preprint arXiv:2403.03949}, 2024.

\bibitem{hu2024privileged}
E.~S. Hu, J.~Springer, O.~Rybkin, and D.~Jayaraman, ``Privileged sensing scaffolds reinforcement learning,'' \emph{arXiv preprint arXiv:2405.14853}, 2024.

\bibitem{hafner2023mastering}
D.~Hafner, J.~Pasukonis, J.~Ba, and T.~Lillicrap, ``Mastering diverse domains through world models,'' \emph{arXiv preprint arXiv:2301.04104}, 2023.

\bibitem{miki2022learning}
T.~Miki, J.~Lee, J.~Hwangbo, L.~Wellhausen, V.~Koltun, and M.~Hutter, ``Learning robust perceptive locomotion for quadrupedal robots in the wild,'' \emph{Science robotics}, vol.~7, no.~62, p. eabk2822, 2022.

\bibitem{lee2020learning}
J.~Lee, J.~Hwangbo, L.~Wellhausen, V.~Koltun, and M.~Hutter, ``Learning quadrupedal locomotion over challenging terrain,'' \emph{Science robotics}, vol.~5, no.~47, p. eabc5986, 2020.

\bibitem{ankile2024imitation}
L.~Ankile, A.~Simeonov, I.~Shenfeld, M.~Torne, and P.~Agrawal, ``From imitation to refinement--residual rl for precise visual assembly,'' \emph{arXiv preprint arXiv:2407.16677}, 2024.

\bibitem{uppal2024spin}
S.~Uppal, A.~Agarwal, H.~Xiong, K.~Shaw, and D.~Pathak, ``Spin: Simultaneous perception interaction and navigation,'' in \emph{Proceedings of the IEEE/CVF Conference on Computer Vision and Pattern Recognition}, 2024, pp. 18\,133--18\,142.

\bibitem{warrington2021robust}
A.~Warrington, J.~W. Lavington, A.~Scibior, M.~Schmidt, and F.~Wood, ``Robust asymmetric learning in pomdps,'' in \emph{International Conference on Machine Learning}.\hskip 1em plus 0.5em minus 0.4em\relax PMLR, 2021, pp. 11\,013--11\,023.

\bibitem{messikommer2024student}
N.~Messikommer, J.~Xing, E.~Aljalbout, and D.~Scaramuzza, ``Student-informed teacher training,'' \emph{arXiv preprint arXiv:2412.09149}, 2024.

\bibitem{walsman2022impossibly}
A.~Walsman, M.~Zhang, S.~Choudhury, D.~Fox, and A.~Farhadi, ``Impossibly good experts and how to follow them,'' in \emph{The Eleventh International Conference on Learning Representations}, 2022.

\bibitem{weihs2021bridging}
L.~Weihs, U.~Jain, I.-J. Liu, J.~Salvador, S.~Lazebnik, A.~Kembhavi, and A.~Schwing, ``Bridging the imitation gap by adaptive insubordination,'' \emph{Advances in Neural Information Processing Systems}, vol.~34, pp. 19\,134--19\,146, 2021.

\bibitem{bellemare2017distributional}
M.~G. Bellemare, W.~Dabney, and R.~Munos, ``A distributional perspective on reinforcement learning,'' in \emph{International conference on machine learning}.\hskip 1em plus 0.5em minus 0.4em\relax PMLR, 2017, pp. 449--458.

\bibitem{ross2011reduction}
S.~Ross, G.~Gordon, and D.~Bagnell, ``A reduction of imitation learning and structured prediction to no-regret online learning,'' in \emph{Proceedings of the fourteenth international conference on artificial intelligence and statistics}.\hskip 1em plus 0.5em minus 0.4em\relax JMLR Workshop and Conference Proceedings, 2011, pp. 627--635.

\bibitem{hallak2015contextual}
A.~Hallak, D.~Di~Castro, and S.~Mannor, ``Contextual markov decision processes,'' \emph{arXiv preprint arXiv:1502.02259}, 2015.

\bibitem{kakade2002approximately}
S.~Kakade and J.~Langford, ``Approximately optimal approximate reinforcement learning,'' in \emph{Proceedings of the nineteenth international conference on machine learning}, 2002, pp. 267--274.

\bibitem{florensa2017reverse}
C.~Florensa, D.~Held, M.~Wulfmeier, M.~Zhang, and P.~Abbeel, ``Reverse curriculum generation for reinforcement learning,'' in \emph{Conference on robot learning}.\hskip 1em plus 0.5em minus 0.4em\relax PMLR, 2017, pp. 482--495.

\bibitem{salimans2018learning}
T.~Salimans and R.~Chen, ``Learning montezuma's revenge from a single demonstration,'' \emph{arXiv preprint arXiv:1812.03381}, 2018.

\bibitem{chen2020learning}
D.~Chen, B.~Zhou, V.~Koltun, and P.~Kr{\"a}henb{\"u}hl, ``Learning by cheating,'' in \emph{Conference on robot learning}.\hskip 1em plus 0.5em minus 0.4em\relax PMLR, 2020, pp. 66--75.

\bibitem{chen2023visual}
T.~Chen, M.~Tippur, S.~Wu, V.~Kumar, E.~Adelson, and P.~Agrawal, ``Visual dexterity: In-hand reorientation of novel and complex object shapes,'' \emph{Science Robotics}, vol.~8, no.~84, p. eadc9244, 2023.

\bibitem{qi2023hand}
H.~Qi, A.~Kumar, R.~Calandra, Y.~Ma, and J.~Malik, ``In-hand object rotation via rapid motor adaptation,'' in \emph{Conference on Robot Learning}.\hskip 1em plus 0.5em minus 0.4em\relax PMLR, 2023, pp. 1722--1732.

\bibitem{liu2024visual}
M.~Liu, Z.~Chen, X.~Cheng, Y.~Ji, R.~Yang, and X.~Wang, ``Visual whole-body control for legged loco-manipulation,'' \emph{arXiv preprint arXiv:2403.16967}, 2024.

\bibitem{choudhury2024better}
S.~Choudhury and P.~Sodhi, ``Better than your teacher: Llm agents that learn from privileged ai feedback,'' \emph{arXiv preprint arXiv:2410.05434}, 2024.

\bibitem{kaelbling1998planning}
L.~P. Kaelbling, M.~L. Littman, and A.~R. Cassandra, ``Planning and acting in partially observable stochastic domains,'' \emph{Artificial intelligence}, vol. 101, no. 1-2, pp. 99--134, 1998.

\bibitem{swamy2022sequence}
G.~Swamy, S.~Choudhury, J.~Bagnell, and S.~Z. Wu, ``Sequence model imitation learning with unobserved contexts,'' \emph{Advances in Neural Information Processing Systems}, vol.~35, pp. 17\,665--17\,676, 2022.

\bibitem{spencer2020learning}
J.~Spencer, S.~Choudhury, M.~Barnes, M.~Schmittle, M.~Chiang, P.~Ramadge, and S.~Srinivasa, ``Learning from interventions: Human-robot interaction as both explicit and implicit feedback,'' in \emph{16th robotics: science and systems, RSS 2020}.\hskip 1em plus 0.5em minus 0.4em\relax MIT Press Journals, 2020.

\bibitem{tavakoli2018exploring}
A.~Tavakoli, V.~Levdik, R.~Islam, C.~M. Smith, and P.~Kormushev, ``Exploring restart distributions,'' \emph{arXiv preprint arXiv:1811.11298}, 2018.

\bibitem{khandate2023sampling}
G.~Khandate, S.~Shang, E.~T. Chang, T.~L. Saidi, Y.~Liu, S.~M. Dennis, J.~Adams, and M.~Ciocarlie, ``Sampling-based exploration for reinforcement learning of dexterous manipulation,'' \emph{arXiv preprint arXiv:2303.03486}, 2023.

\end{thebibliography}

\addtolength{\textheight}{-12cm}

\section*{Appendix}
\subsection{Critical State Selection}
\label{app:critical_state}
A \textit{critical state} is a state where querying the expert most effectively reduces imitation loss in the next training iteration. 
Given a well-trained student policy with sufficiently small model capacity error, i.e.,
\begin{equation}
    \epsilon = \mathbb{E}_{s \sim D(s)} \Big[ D(\pi(s) \| \pi^*(s)) \Big] < \alpha,
\end{equation}
where \( \alpha \) is a small threshold ensuring the student's policy deviation from the expert remains controlled, the critical state selection follows a two-step process:

\paragraph{Transition Outside Dataset Condition (Necessary Condition)}
A state \( s \) is considered a candidate for being a critical state if executing the student policy \( \pi(s) \) results in a next state \( s' \) that is not present in the dataset:
\begin{equation}
    P_c[s' | s, \pi(s)], \quad s' \notin \mathcal{D}(s).
\end{equation}
This ensures that the student is about to enter an unseen region where expert supervision is necessary. Without this condition, any state along an incorrect trajectory could be selected, even if the student could still recover.

\paragraph{Optimization Criterion for Selecting the Best Critical State}
Among states that satisfy the transition condition, we select the \textit{best} critical state by minimizing the trade-off between: state aliasing, useful data augmentation and state connectivity.

\begin{equation}
    s_\text{critical} = \arg \min_{s\in\mathcal{S}_N} \Big( (\delta_T - \delta_{t_\text{critical}}) + \lambda D(s, \mathcal{D}) \Big),
\end{equation}

where:
\begin{itemize}
    \item \( \delta_T - \delta_{t_\text{critical}} \) ensures that the expert trajectory balances:
    \begin{itemize}
        \item Minimizing aliasing impact: Preventing excessive aliasing when incorporating new data.
        \item Maximizing useful data augmentation: Ensuring that the additional expert trajectory contributes to cover new feasible state manifold that wasn't covered at previous dataset.
    \end{itemize}
    \item \( D(s, \mathcal{D}) \) measures the connectivity of state \( s \) to the expert dataset \( \mathcal{D} \), ensuring that the selected critical state remains within a distribution where expert intervention is meaningful.
\end{itemize}

Once the critical state is chosen, the expert trajectory \( \mathbf{\tau}_i \) collected during the \( i \)-th augmentation iteration, starting from \( s_\text{critical} \) at time \( t \) to the final state \( s_T \), is aggregated into the dataset \( \mathcal{D}_{i-1} \):
\begin{equation}
    \mathbf{\tau}_i = \{s_\text{critical}, a_t, s_{t+1}, a_{t+1}, \dots, s_T\}_i.
\end{equation}
\begin{equation}
    \mathcal{D}_{i} \leftarrow \mathcal{D}_{i-1} \cup \mathbf{\tau}_i.
\end{equation}
The training and data augmentation process is then iteratively repeated with the updated dataset.

\end{document}